\documentclass{article}

\usepackage{microtype}
\usepackage{graphicx}
\usepackage{subfigure}
\usepackage{booktabs} 

\usepackage[colorlinks=true, linkcolor=black, citecolor=black, urlcolor=blue!80!black]{hyperref}

\usepackage[accepted]{icml2025}

\usepackage{amsmath}
\usepackage{amssymb}
\usepackage{mathtools}
\usepackage{amsthm}

\usepackage[capitalize,noabbrev]{cleveref}

\usepackage{enumitem}
\usepackage{amsmath}
\usepackage{amssymb}
\usepackage{booktabs}
\usepackage{makecell}
\usepackage{threeparttable}
\usepackage{array}
\usepackage{afterpage}
\usepackage{graphicx}

\newcommand{\ieno}{\textit{i.e.}}
\newcommand{\egno}{\textit{e.g.}}
\theoremstyle{plain}

\theoremstyle{definition}

\theoremstyle{remark}

\usepackage[textsize=tiny]{todonotes}

\begin{document}

\twocolumn[
\icmltitle{QMamba: On First Exploration of Vision Mamba for Image Quality Assessment}




\begin{icmlauthorlist}
\icmlauthor{Fengbin Guan}{ustc}
\icmlauthor{Xin Li}{ustc}
\icmlauthor{Zihao Yu}{ustc}
\icmlauthor{Yiting Lu}{ustc}
\icmlauthor{Zhibo Chen}{ustc}\\

\texttt{\small \{guanfb, yuzihao, luyt31415\}@mail.ustc.edu.cn, \{xin.li, chenzhibo\}@ustc.edu.cn}

\end{icmlauthorlist}
\icmlaffiliation{ustc}{University of Science and Technology of China, Hefei, China}

\icmlcorrespondingauthor{Xin Li}{xin.li@ustc.edu.cn}

\icmlkeywords{Machine Learning, ICML}

\vskip 0.3in
]



\printAffiliationsAndNotice{}  

\begin{abstract}
In this work, we take the first exploration of the recently popular foundation model, \ieno, State Space Model/Mamba, in image quality assessment (IQA), aiming at observing and excavating the perception potential in vision Mamba. 
A series of works on Mamba has shown its significant potential in various fields, \egno, segmentation and classification. However, the perception capability of Mamba remains under-explored. Consequently, we propose QMamba by revisiting and adapting the Mamba model for three crucial IQA tasks, \ieno, task-specific, universal, and transferable IQA, which reveals its clear advantages over existing foundational models, \egno, Swin Transformer, ViT, and CNNs, in terms of perception and computational cost. To improve the transferability of QMamba, we propose the StylePrompt tuning paradigm, where lightweight mean and variance prompts are injected to assist task-adaptive transfer learning of pre-trained QMamba for different downstream IQA tasks. Compared with existing prompt tuning strategies, our StylePrompt enables better perceptual transfer with lower computational cost. Extensive experiments on multiple synthetic, authentic IQA datasets, and cross IQA datasets demonstrate the effectiveness of our proposed QMamba. The code will be available at: \href{https://github.com/bingo-G/QMamba.git}{https://github.com/bingo-G/QMamba.git}
\end{abstract}

\begin{figure}
    \centering
    \includegraphics[width=1\linewidth]{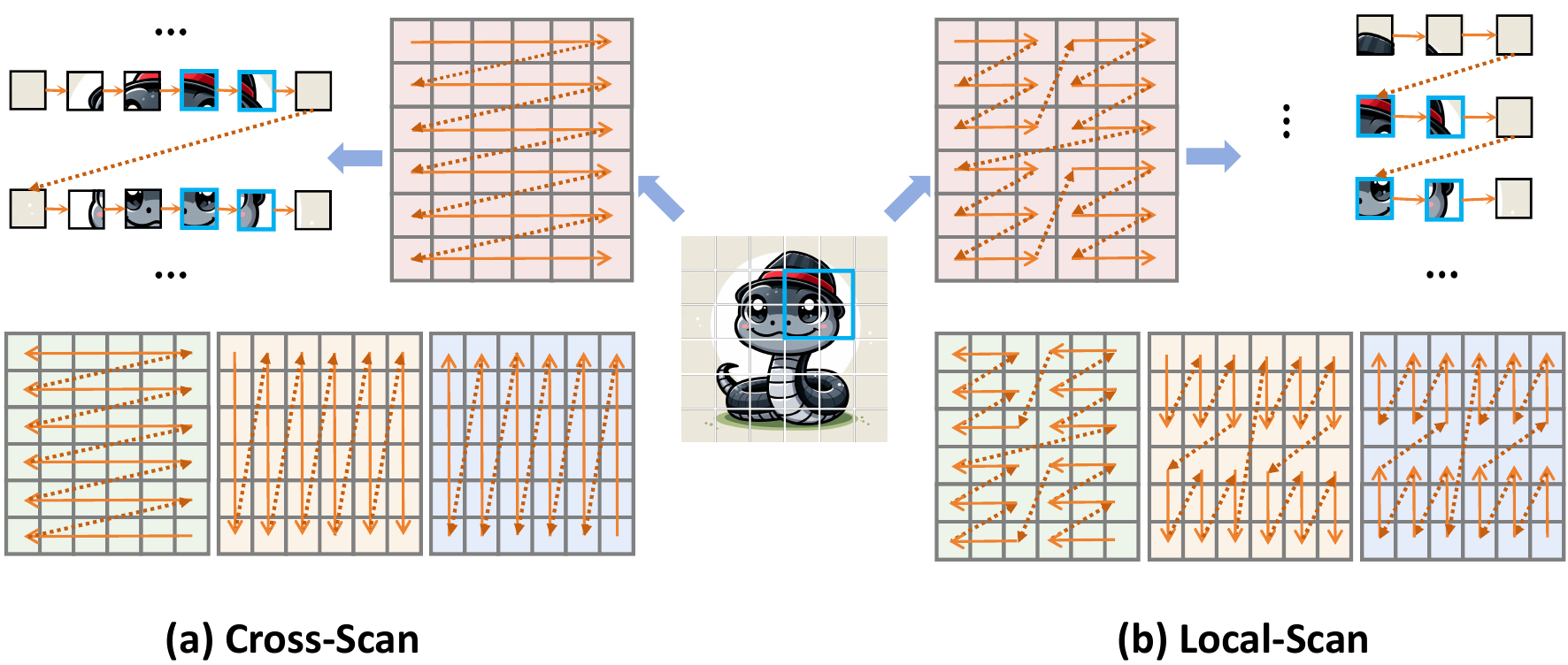}
    \caption{Scanning Methodology Illustration. (a) VMamba's scanning method\cite {VMamba} flattens 2D data into 1D, impairing connectivity by distancing adjacent tokens. (b) Local scanning method scans within and across windows, placing semantically similar and distortion-related tokens closer, as shown in the blue boxes.}

    \label{fig: Localscan}
\end{figure}

\section{Introduction}
\label{submission}
Image Quality Assessment (IQA) aims to measure the subjective quality of images aligned with human perception, which has been applied in various visual fields, including visual acquisition, transmission, AIGC~\cite{AGIQA3K, AIGCIQA2023}, and UGC creation~\cite{UGC, lu2024kvq}, etc. Establishing a great IQA metric is necessary to provide the right optimization direction tailored for image processing techniques, \ieno, compression~\cite{compression,compression2,swinqadcc}, enhancement~\cite{li2023learningDIL, fei2023generative}, and ensure the perceptual quality of images. Early works on IQA have been achieved by leveraging the natural scene statistics in a hand-crafted manner~\cite{hand1, hand2}. With the advancements of deep neural networks (DNNs), learning-based IQA metrics~\cite{li2023freqalign,liu2022liqa} have demonstrated significant potential for low-level perception, which can be roughly categorized into two types based on the pre-trained backbones: CNN-based and Transformer-based methods.

Although the impressive progress, learning-based IQA is susceptible to inherent limitations of existing pre-trained backbones: (i) the CNNs are skilled at learning local translation-invariant features from images while lacking enough long-range dependency modeling capability, hindering the global quality perception. (ii) The emergence of Vision Transformers presents a great solution to model long-range dependency effectively by leveraging attention mechanisms. However, the quadratic complexity of self-attention operations poses unaffordable computational costs, especially for large-scale image quality assessment. Recently, an innovative foundation model, the State Space Model, particularly its implementation, \ieno, Mamba~\cite{Mamba} has shown considerable potential in various fields for balancing the computational costs and performances, \egno, segmentation\cite{Segmamba, Segment-U-mamba} and classification\cite{VMamba, Vim}. This raises one interesting question, \ieno, ``whether Mamba can surpass existing backbones on low-level visual perception", which is still under-explored.

To answer this question, in this work, we initiate a new exploration of the Mamba model within the field of IQA and introduce the QMamba, a newly developed IQA framework designed to address three key facets of IQA: task-specific, universal, and transferable image quality assessment. Notably, since the lack of enough IQA dataset, learning-based IQA metrics entail the pre-trained backbone for perception knowledge extraction. Meanwhile,  VMamba\cite{VMamba}, as the most representative framework, has achieved excellent performance on high-level tasks by employing horizontal and vertical scanning strategies. However, merely excavating the global perception knowledge is not optimal for IQA, since most artifacts affecting the image quality are related to local textures. Consequently, inspired by LocalMamba\cite{Localmamba}, we have adopted a local window scanning method, which significantly enhances ability of Mamba to perceive local distortions, thereby demonstrating superior performance. We detail the architecture of QMamba and examine different scanning methods to clarify how the model interacts with and processes image data. Our extensive analysis confirms that QMamba surpasses traditional foundational models, showing superior perceptual accuracy and greater computational efficiency.

Moreover, we have explored the perceptual transferability of Mamba across different datasets, \ieno, different contents, and degradations. We can find that the Mamba-based IQA metric still suffers from severe performance drop when they encounter large domain shifts between synthetic distortions\cite{live2, Kadid}, authentic distortions\cite{livec, Koniq}, and Artificial Intelligence-Generated Content (AIGC)\cite{AGIQA3K, AIGCIQA2023} distortions, etc. To further amplify the transferability of QMamba across a range of IQA applications, we introduce a simple but effective tuning strategy called StylePrompt. This is based on the finding that the domain shifts in IQA tend to correlate with their feature statistics/style~\cite{styleam}, such as mean and variance. Concretely, our StylePrompt aims to adaptively adjust the mean and variance of pre-trained QMamba towards the target IQA tasks by setting a group of light-weight learnable $1\times 1\times C$ parameters. Extensive experiments have shown that our StylePrompt greatly facilitates the task-adaptive learning of QMamba, enabling efficient knowledge transfer across diverse IQA tasks with fewer parameter costs.

The main contributions of this paper are summarized as follows:
\begin{itemize}

   \item We embark on a novel exploration of the Mamba model within image quality assessment, and propose the QMamba, a powerful IQA metric for 
   three critical tasks of IQA: task-specific, universal, and transferable image quality assessment. This exploration has demonstrated the superior potential of Mamba for subjective perception, advancing the development of IQA.

    \item To improve the perception transferability of the QMamba, we introduce a simple but effective tuning strategy, \ieno, StylePrompt. This strategy enables the efficient knowledge transfer of pre-trained QMamba for downstream IQA tasks, while only tuning fewer learnable parameters to adjust the statistics of perception features.

    \item Extensive experiments have shown that our QMamba has consistently achieved state-of-the-art results on various prominent IQA datasets compared with existing IQA methods, thereby validating the efficacy of the Mamba model in quality assessment. Moreover, our StylePrompt achieves nearly equivalent performance to full model tuning while utilizing only $4\%$ of the whole parameters, demonstrating its effectiveness in perception knowledge transfer scenarios.
    
\end{itemize}

\section{Related Work}

\subsection{Blind Image Quality Assessment (BIQA)}

Early BIQA methods relied heavily on manually designed features for quality score regression \cite{hand1, hand2, hand3, hand4,BMPRI}. However, these handcrafted features were insufficient for addressing the complexity of BIQA tasks. With the advent of deep learning, network architectures capable of powerful feature extraction significantly improved quality assessment tasks, with Convolutional Neural Networks (CNNs) and Vision Transformers being the most prevalent.

\textbf{CNN-based BIQA.} CNNs have demonstrated robust feature extraction capabilities, leading to their widespread adoption in BIQA tasks. Early works like CNNIQA \cite{CNNIQA} used convolutional models for feature learning and quality regression, substantially outperforming handcrafted features. DBCNN \cite{DBCNN} introduced a dual-stream network to address synthetic and authentic distortions separately, integrating these insights for better quality prediction. NIMA \cite{NIMA} and PQR \cite{PQR} leveraged pre-trained models on ImageNet for quality score prediction, enhancing accuracy through well-established neural architectures. MetaIQA \cite{metaiqa} employed meta-learning to adapt to unknown distortions by learning shared priors for various distortion types, while HyperIQA \cite{HyperIQA} used a hypernetwork to adaptively establish perceptual rules, improving generalization. Despite these advancements\cite{ReIQA,QPT,TOPIQ}, CNNs' local bias limits their ability to fully exploit both global and local information in BIQA tasks.

\begin{figure*}
    \centering
    \includegraphics[width=0.95\linewidth]{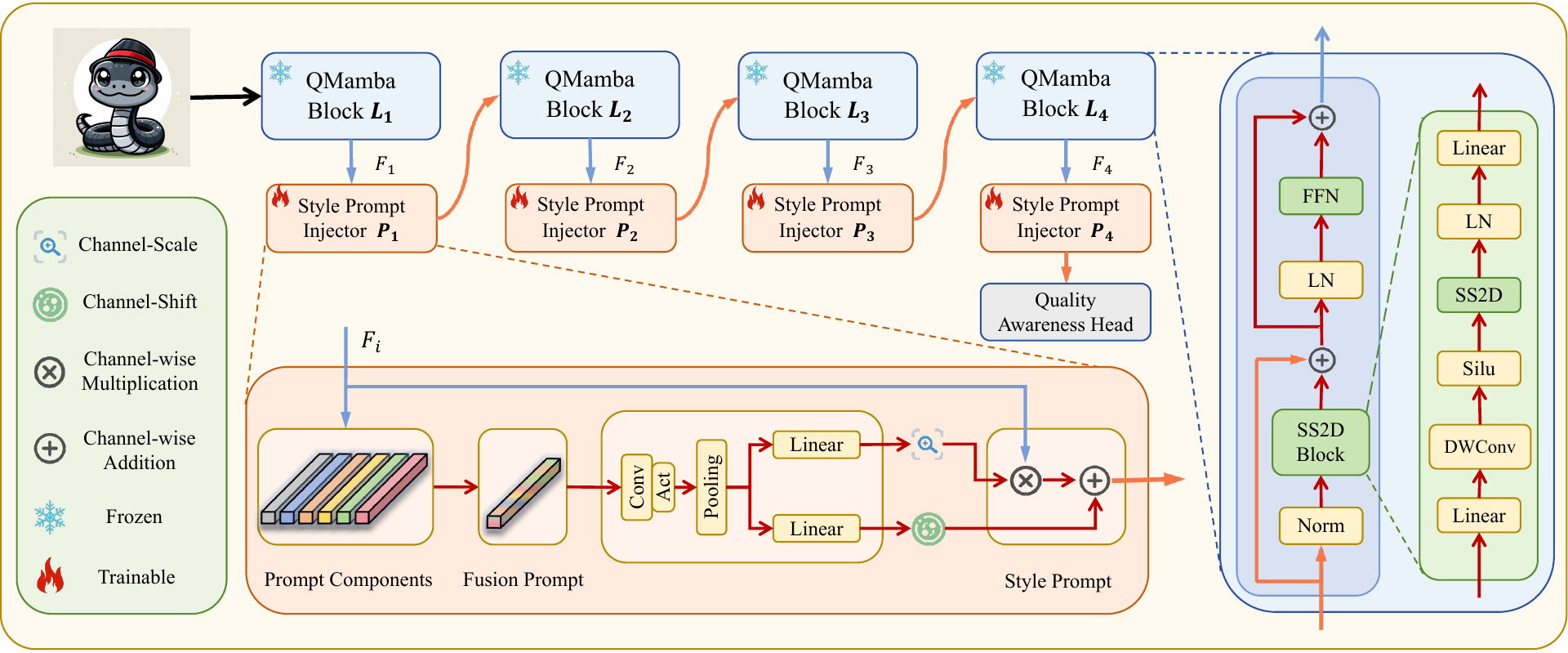}
    \caption{Architectural Overview of QMamba Framework and the Detailed of StylePrompt Tuning Mechanism.}
    \label{fig: QMamba and StylePrompt}
\end{figure*}

\textbf{Transformer-based BIQA.} Transformers offer superior global modeling capabilities compared to CNNs. TReS \cite{TReS} addressed CNNs' local bias by capturing local structural information with CNNs and then using Transformers for sequential feature extraction. MUSIQ \cite{musiq} designed a multi-scale image Transformer architecture capable of handling images with varying sizes and aspect ratios. DEIQT \cite{DEIQT} leveraged a Transformer-based BIQA architecture with attention mechanisms to align with human perception, enhancing model performance and reducing prediction uncertainty. However, the quadratic complexity of Transformers presents a challenge, highlighting the need for architectures with linear complexity capable of global modeling for BIQA tasks. In addition to the above methods, many recent works \cite{LIQE,QCN,sf-IQA} have also adopted Transformer-based designs and achieved competitive performance, yet they still face limitations due to high computational complexity.

\subsection{State Space Models (SSMs)}

State space models, known for their linear complexity in capturing long-range dependencies, have been integrated into deep learning architectures. The Structured State Space model (S4) \cite{S4} was a pioneer in deep state space modeling for remote dependency modeling. Subsequent advancements \cite{Hippo, S5, H3} further propelled the development of state space models. Mamba \cite{Mamba}, by integrating selection mechanisms and hardware-aware algorithms, has shown effective long-range modeling capabilities with linear complexity growth.

Initially focused on NLP tasks, Mamba has rapidly expanded into other domains. Vim \cite{Vim} introduced a bidirectional SSM block for visual representation learning, achieving performance comparable to ViT \cite{ViT}. VMamba \cite{VMamba} introduced a cross-scan module to traverse spatial domains and convert non-causal visual images into ordered block sequences, maintaining linear complexity while retaining global receptive fields. LocalMamba \cite{Localmamba} employed a window-based scanning approach to integrate local inductive biases, enhancing the visual Mamba model. These advancements validate the efficacy of Mamba in visual tasks, leading to its application in image classification \cite{VMamba, Vim, Simba}, video understanding \cite{MambaVideoUnderstanding, Videomambasuite, videomamba}, image restoration \cite{MambaIR, VmambaIR,zhen2024freqmamba}, point cloud analysis \cite{pointmamba, point,liu2024point}, and biomedical image segmentation \cite{Segment-U-mamba, Segmamba, Swin-umamba}. These studies have demonstrated the effectiveness of state space models in visual tasks, providing a solid foundation for further exploration of their potential in visual perception.

\section{Method}
\label{method}

\subsection{Exploring Mamba for Perception}
\subsubsection{Overall Framework}

The overall architecture of our proposed QMamba model is depicted in Figure~\ref{fig: QMamba and StylePrompt}. To address the unique challenges of visual perception tasks, we developed a novel architecture that rethinks the design principles of state space models for visual data. The network incorporates a hierarchical residual structure, where convolutional layers enable effective feature extraction while specialized activation layers compute adaptive gating signals. Our architecture organizes the processing into multiple network stages, each combining a strategic downsampling layer with our enhanced Mamba-based processing block. This design enables the construction of multi-level representations at varying resolutions, facilitating the extraction of richer perceptual features through progressive abstraction.

To systematically investigate the relationship between model capacity and quality perception performance, we developed three distinct variants of our architecture: QMamba-Tiny, QMamba-Small, and QMamba-Base. Each variant maintains the core architectural principles while scaling in complexity, allowing us to explore the trade-offs between computational efficiency and perceptual accuracy across different application scenarios.

\subsubsection{Perception with Local Scanning}

While the original Mamba model excels in natural language processing tasks with inherently causal inputs, it faces challenges when applied to visual tasks due to the absence of spatial causality. In particular, it struggles to capture complex spatial dependencies among image pixels, which hinders its ability to model local distortions effectively.

To address this, VMamba~\cite{VMamba} introduces a bidirectional horizontal and vertical scanning strategy to convert 2D images into 1D sequences suitable for sequential modeling. Although this enables global pixel-level modeling, it disrupts the continuity of locally adjacent tokens, weakening the model’s ability to perceive fine-grained distortions—an essential factor for image quality assessment.

Inspired by LocalMamba~\cite{Localmamba}, we adopt a window-based scanning approach that performs horizontal scans within local windows, followed by window-level scans, and applies the same strategy vertically. This method enhances local distortion perception while retaining global awareness through hierarchical composition. Despite downsampling and aggregation in the input, this scanning scheme maintains a balance between local detail and broader context.

Unlike LocalMamba, which relies on attention-based dynamic routing and suffers from unstable inference and high computational cost, LQMamba adopts a hierarchical architecture with fixed-size windows that change with network depth. As illustrated in Figure~\ref{fig: Localscan} (b), this structure enables the model to capture multi-scale perceptual cues, from fine-grained distortions to broader contextual patterns, while maintaining stable and efficient inference. It achieves a good balance between accuracy and efficiency, making it well-suited for IQA tasks that demand consistent perception across various distortion types and spatial scales.

\begin{figure*}
    \centering
    \includegraphics[width=0.97\linewidth]{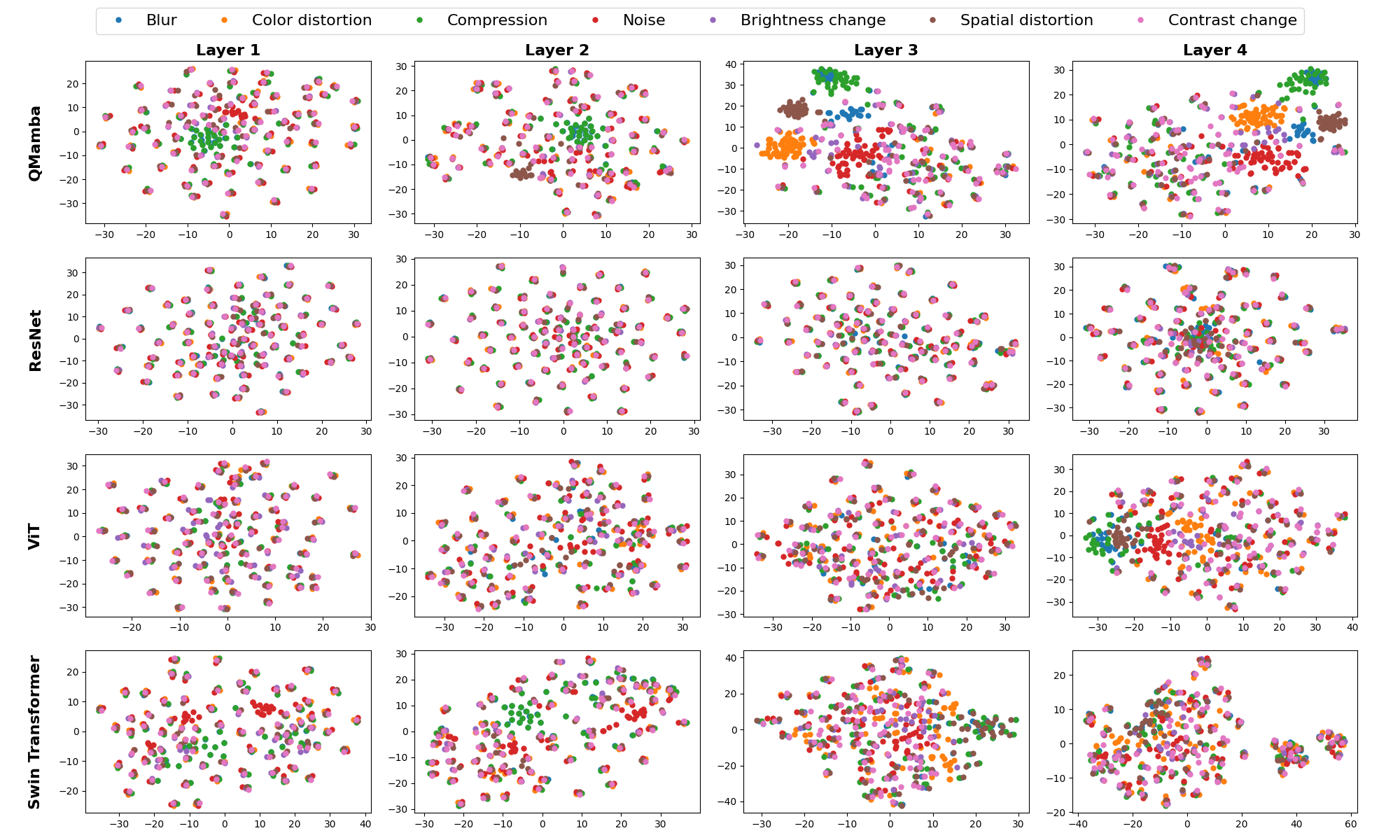}
    \caption{t-SNE Visualization of Distortion-Specific Feature Separation: QMamba vs. Conventional Backbones}

    \label{fig: t-SNE}
\end{figure*}

\subsubsection{Analysis of Perceptual Capability in State Space Models}

To investigate the feature selection mechanism of State Space Models (\ieno ,Mamba) in visual quality assessment tasks, we employ t-SNE visualization for deep-layer feature analysis, as illustrated in Figure \ref{fig: t-SNE}. Visualization results reveal that Mamba exhibits distinctive characteristic evolution patterns through dynamic state updating mechanisms, progressively enhancing distortion semantic perception across network hierarchies. Specifically, shallow layers preserve diverse features of original visual signals, while deeper layers adaptively filter redundant background information through gated state selection, intensifying distortion-type-specific feature focusing. This drives distortion-homogeneous samples to form cluster structures in feature space while amplifying inter-class discriminability. Such feature refinement strategy not only retains critical discriminative information for quality assessment but also significantly improves model sensitivity to quality degradation cues, thereby strengthening prediction robustness.

\subsection{Tuning the Mamba with StylePrompt}

Although the Mamba architecture reduces computational complexity compared to other models, achieving higher performance still requires a substantial number of parameters, which poses a challenge for efficient transferable learning, crucial for IQA tasks. We observe that domain shifts in IQA tasks tend to correlate with their feature statistics or style~\cite{styleam}, such as mean and variance. Building on this insight, we propose a lightweight tuning strategy, StylePrompt, designed to adjust the mean and variance of the pre-trained QMamba features, thereby aligning them with the distortions and content types of the target domain. This approach enables us to achieve results comparable to full-parameter fine-tuning while using a minimal number of parameters, significantly enhancing both the efficiency and performance of QMamba in transferable IQA tasks. Figure~\ref{fig: QMamba and StylePrompt} illustrates the StylePrompt, which consists of two components that will be described in detail below:

\subsubsection{StylePrompt Generation (SPG)}

We designed the StylePrompt Generation phase to facilitate the creation of prompts and their interaction with the original features. In a multi-stage network architecture, as images progress through each stage of the network, we learn a set of prompts \( P_s \in \mathbb{R}^{N\!\times\!1\!\times\!1\!\times\!C} \), containing N prompt components designed to generate affine parameters for style adaptation. These prompts are specifically utilized to inject the distortion information of the current data into the features \( F_i \in \mathbb{R}^{\hat{H} \times \hat{W} \times \hat{C}} \), facilitating learning of the style pertinent to the target domain.

To enable the prompt components to extract specific style information from the input stream dynamically, we predict the weights for different prompt components based on the input features. This process involves performing a global pooling on the current layer’s features followed by applying a softmax function to obtain the weights for the prompt group. These weights are then applied to the multiple prompt components to amalgamate them into a new prompt \( P_f \), effectively encapsulating the current style information. The operation can be briefly summarized by the following formula:
\begin{equation}
P_f = \sum_{c=1}^{N} w_s P_s , \quad w_s = \text{Softmax}(\text{Conv1x1}(\text{GAP}(F_i)))
\end{equation}

\subsubsection{StylePrompt Injection (SPI)}

In the SPG phase, the fused prompt \( P_f \) is created, containing the distortion style information of the target domain. During the subsequent StylePrompt Injection process, this style information is injected into the original features \( F_i \in \mathbb{R}^{\hat{H} \times \hat{W} \times \hat{C}} \) produced by the current layer. To achieve this, \( P_f \) is matched to the channel dimensions of the current features using linear layers designed for dimensional alignment. The processed prompt then generates affine parameters \( \gamma_v \in \mathbb{R}^{1 \times 1 \times \hat{C}} \) and \( \beta_v \in \mathbb{R}^{1 \times 1 \times \hat{C}} \), which modulate the mean and variance of the original feature distribution solely along the channel dimension. This adaptive adjustment ensures that the feature distribution is tailored to the distortion style of the target domain with minimal computational overhead.

 The process of our StylePrompt Injection can be summarized as follows:
\begin{equation}
\gamma_v = \text{Linear}_{\gamma}(\text{Conv}(P_f))
\end{equation}
\begin{equation}
\beta_v = \text{Linear}_{\beta}(\text{Conv}(P_f))
\end{equation}
\begin{equation}
F{'}_i = F_i \cdot (1 + \gamma_v) + \beta_v
\end{equation}

Specifically, \( F_i' \) represents the original features after the injection of style information. These enhanced features will serve as the new input to the subsequent layer of the network.

\begin{table*}[h!]
\centering

\renewcommand{\arraystretch}{1.3}
\resizebox{\linewidth}{!}{
\setlength{\tabcolsep}{0.8mm}{
\begin{threeparttable}
\begin{tabular}{cccccccccc|ccccccccc}
\Xhline{1.2pt}
                                  &                             & \multicolumn{2}{c}{LIVE}        & \multicolumn{2}{c}{CSIQ}        & \multicolumn{2}{c}{TID2013}     & \multicolumn{2}{c|}{KADID}      & \multicolumn{2}{c}{LIVEC}       & \multicolumn{2}{c}{KonIQ}       & \multicolumn{2}{c}{LIVEFB}      & \multicolumn{2}{c}{SPAQ}        &                \\ \Xhline{1pt}
\multicolumn{1}{c|}{Method}                   & \multicolumn{1}{c|}{GFLOPS}  & PLCC           & SRCC           & PLCC           & SRCC           & PLCC           & SRCC           & PLCC           & SRCC           & PLCC           & SRCC           & PLCC           & SRCC           & PLCC           & SRCC           & PLCC           & SRCC           & Average        \\ \Xhline{1pt}
\multicolumn{1}{c|}{ILNIQE}             & \multicolumn{1}{c|}{-}      & 0.906          & 0.902          & 0.865          & 0.822          & 0.648          & 0.521          & 0.558          & 0.534          & 0.508          & 0.508          & 0.537          & 0.523          & 0.332          & 0.294          & 0.712          & 0.713          & 0.618          \\
\multicolumn{1}{c|}{BRISQUE}            & \multicolumn{1}{c|}{-}      & 0.944          & 0.929          & 0.748          & 0.812          & 0.571          & 0.626          & 0.567          & 0.528          & 0.629          & 0.629          & 0.685          & 0.681          & 0.341          & 0.303          & 0.817          & 0.809          & 0.664          \\
\multicolumn{1}{c|}{WaDIQaM(5.24M)}            & \multicolumn{1}{c|}{2.43G}      & 0.955          & 0.960          & 0.844          & 0.852          & 0.855          & 0.835          & 0.752          & 0.739          & 0.671          & 0.682          & 0.807          & 0.804          & 0.467          & 0.455          & -              & -              & 0.763          \\
\multicolumn{1}{c|}{DBCNN(15.31M)}              & \multicolumn{1}{c|}{16.51G}      & 0.971          & 0.968          & \textbf{0.959} & \textbf{0.946} & 0.865          & 0.816          & 0.856          & 0.851          & 0.869          & 0.851          & 0.884          & 0.875          & 0.551          & 0.545          & 0.915          & 0.911          & 0.852          \\
\multicolumn{1}{c|}{TIQA(23.68M)}               & \multicolumn{1}{c|}{-}      & 0.965          & 0.949          & 0.838          & 0.825          & 0.858          & 0.846          & 0.855          & 0.850          & 0.861          & 0.845          & 0.903          & 0.892          & 0.581          & 0.541          & -              & -              & 0.829          \\
\multicolumn{1}{c|}{MetaIQA(13.24M)}            & \multicolumn{1}{c|}{1.82G}      & 0.959          & 0.960          & 0.908          & 0.899          & 0.868          & 0.856          & 0.775          & 0.762          & 0.802          & 0.835          & 0.856          & 0.887          & 0.507          & 0.540          & -              & -              & 0.815          \\
\multicolumn{1}{c|}{HyperIQA(27.38M)}           & \multicolumn{1}{c|}{4.31G}      & 0.966          & 0.962          & 0.942          & 0.923          & 0.858          & 0.840          & 0.845          & 0.852          & 0.882          & 0.859          & 0.917          & 0.906          & 0.602          & 0.544          & 0.915          & 0.911          & 0.858          \\
\multicolumn{1}{c|}{TReS(152.45M)}               & \multicolumn{1}{c|}{20.03G}      & 0.968          & 0.969          & 0.942          & 0.922          & 0.883          & 0.863          & 0.858          & 0.859          & 0.877          & 0.846          & 0.928          & 0.915          & 0.625          & 0.554          & -              & -              & 0.858          \\
\multicolumn{1}{c|}{MUSIQ(27.13M)}              & \multicolumn{1}{c|}{9.02G}      & 0.911          & 0.940          & 0.893          & 0.871          & 0.815          & 0.773          & 0.872          & 0.875          & 0.746          & 0.702          & 0.928          & 0.916          & 0.661          & 0.566          & 0.921          & 0.918          & 0.832          \\
\multicolumn{1}{c|}{DEIQT(24.04M)}              & \multicolumn{1}{c|}{5.41G}      & \textbf{0.982} & \textbf{0.980} & \textbf{0.963} & \textbf{0.946} & 0.908          & 0.892          & 0.887          & 0.889          & 0.894          & 0.875          & 0.934          & 0.921          & 0.663          & 0.571          & 0.923          & 0.919          & 0.884          \\
\multicolumn{1}{c|}{LoDa(*)}               & \multicolumn{1}{c|}{23.74G}      & \textbf{0.979} & \textbf{0.975} & -              & -              & 0.901          & 0.869          & 0.936          & 0.931          & 0.899          & 0.876          & 0.944          & 0.932          & \textbf{0.679} & \textbf{0.578} & 0.928          & 0.925          & 0.882          \\ \Xhline{1pt}
\multicolumn{1}{c|}{ResNet-50(23.51M)}  & \multicolumn{1}{c|}{4.11G}  & 0.879          & 0.884          & 0.861          & 0.841          & 0.747          & 0.686          & 0.784          & 0.786          & 0.868          & 0.831          & 0.908          & 0.886          & 0.313          & 0.269          & 0.907          & 0.907          & 0.772          \\
\multicolumn{1}{c|}{ResNet-101(42.50M)} & \multicolumn{1}{c|}{7.83G}  & 0.918          & 0.921          & 0.891          & 0.867          & 0.779          & 0.727          & 0.722          & 0.719          & 0.862          & 0.824          & 0.918          & 0.904          & 0.420          & 0.347          & 0.908          & 0.906          & 0.790          \\
\multicolumn{1}{c|}{ResNet-152(58.15M)} & \multicolumn{1}{c|}{11.53G} & 0.926          & 0.927          & 0.923          & 0.899          & 0.765          & 0.717          & 0.764          & 0.760          & 0.859          & 0.816          & 0.919          & 0.898          & 0.433          & 0.353          & 0.907          & 0.907          & 0.798          \\ \Xhline{1pt}
\multicolumn{1}{c|}{ViT-T(5.52M)}       & \multicolumn{1}{c|}{1.26G}  & 0.786          & 0.792          & 0.725          & 0.717          & 0.728          & 0.699          & 0.832          & 0.836          & 0.777          & 0.730          & 0.852          & 0.852          & 0.521          & 0.461          & 0.896          & 0.896          & 0.756          \\
\multicolumn{1}{c|}{ViT-S(21.67M)}      & \multicolumn{1}{c|}{4.61G}  & 0.900          & 0.896          & 0.832          & 0.815          & 0.873          & 0.859          & 0.893          & 0.894          & 0.831          & 0.799          & 0.922          & 0.905          & 0.539          & 0.443          & 0.919          & 0.917          & 0.827          \\
\multicolumn{1}{c|}{ViT-B(85.80M)}      & \multicolumn{1}{c|}{17.58G} & 0.961          & 0.955          & 0.924          & 0.912          & 0.904          & 0.905          & 0.910          & 0.908          & 0.875          & 0.837          & 0.913          & 0.895          & 0.491          & 0.452          & 0.914          & 0.912          & 0.854          \\ \Xhline{1pt}
\multicolumn{1}{c|}{Swin-T(27.52M)}     & \multicolumn{1}{c|}{4.51G}  & 0.879          & 0.883          & 0.865          & 0.847          & 0.937          & 0.925          & 0.923          & 0.922          & 0.880          & 0.845          & 0.901          & 0.881          & 0.476          & 0.453          & 0.922          & 0.919          & 0.841          \\
\multicolumn{1}{c|}{Swin-S(48.84M)}     & \multicolumn{1}{c|}{8.77G}  & 0.883          & 0.896          & 0.884          & 0.874          & 0.931          & 0.918          & 0.895          & 0.894          & 0.907          & \textbf{0.884}          & 0.931          & 0.914          & 0.476          & 0.433          & 0.918          & 0.915          & 0.847          \\
\multicolumn{1}{c|}{Swin-B(86.74M)}     & \multicolumn{1}{c|}{15.47G} & 0.945          & 0.948          & 0.941          & 0.935          & 0.942          & 0.933          & 0.934          & 0.932          & 0.892          & 0.858          & 0.945          & 0.932          & 0.507          & 0.471          & 0.923          & 0.921          & 0.872          \\ \Xhline{1pt}
\multicolumn{1}{c|}{QMamba-T (27.99M)}  & \multicolumn{1}{c|}{4.47G}  & 0.959          & 0.959          & 0.940          & 0.918          & 0.951          & 0.945          & 0.934          & 0.930          & 0.898          & 0.866          & 0.941          & 0.925          & 0.675          & 0.581          & \textbf{0.934} & \textbf{0.929} & 0.893          \\
\multicolumn{1}{c|}{QMamba-S (49.37M)}  & \multicolumn{1}{c|}{8.71G}  & 0.962          & 0.965          & 0.921          & 0.903          & \textbf{0.957} & \textbf{0.955} & 0.934          &\textbf{ 0.933}          & 0.903          & 0.874          & 0.943          & 0.930          & \textbf{0.677} & \textbf{0.573} & 0.932          & 0.927          & 0.893          \\
\multicolumn{1}{c|}{QMamba-B (87.53M)}  & \multicolumn{1}{c|}{15.35G} & 0.960          & 0.961          & 0.908          & 0.889          & 0.953          & 0.949          & 0.935          & 0.932          & \textbf{0.908} & 0.876 & 0.943          & 0.930          & 0.675          & 0.579          & 0.933          & 0.929          & 0.891          \\ \Xhline{1pt}
\multicolumn{1}{c|}{LQMamba-T(29.87M)}  & \multicolumn{1}{c|}{4.44G}  & 0.958          & 0.959          & 0.935          & 0.916          & 0.952          & 0.950          & 0.938          & 0.923          & 0.903          & 0.863          & 0.943          & 0.928          & 0.672          & 0.574          & 0.933          & 0.927          & 0.892          \\
\multicolumn{1}{c|}{LQMamba-S(52.91M)}  & \multicolumn{1}{c|}{8.66G}  & 0.962          & 0.964          & 0.933          & 0.914          & 0.955          & 0.949          & \textbf{0.941} & 0.928 & 0.907          & 0.882          & \textbf{0.946} & \textbf{0.934} & 0.676          & 0.574          & 0.933          & 0.929          & \textbf{0.895} \\
\multicolumn{1}{c|}{LQMamba-B(93.79M)}  & \multicolumn{1}{c|}{15.30G} & 0.959          & 0.951          & 0.915          & 0.889          & \textbf{0.965} & \textbf{0.964} & \textbf{0.943} & \textbf{0.941} & \textbf{0.913} & \textbf{0.888} & \textbf{0.947} & \textbf{0.933} & 0.675          & 0.582          & \textbf{0.934} & \textbf{0.929} & \textbf{0.896} \\ \Xhline{1.2pt}
\end{tabular}
\begin{tablenotes}
\item[\large*]   LoDa has a total of 118.23M model parameters and 8.93M trainable parameters.
\end{tablenotes}
\end{threeparttable}
}
}
\caption{Performance Comparison for Task-Specific IQA. Bold Indicates the Top Two Results.}
\label{table: task_specfic_result}
\end{table*}

\section{Experiments}
\subsection{Experimental Setup}
\subsubsection{Datasets}

We conducted foundational experiments on ten popular IQA datasets, which include four synthetic datasets: LIVE\cite{live2}, CSIQ\cite{CSIQ}, TID2013\cite{Tid2013}, and KADID\cite{Kadid}; four authentic datasets: LIVEC\cite{livec}, KonIQ\cite{Koniq}, LIVEFB\cite{LIVEFB}, and SPAQ\cite{SPAQ}; and two AIGC datasets: AIGC2023\cite{AIGCIQA2023} and AGIQA3K\cite{AGIQA3K}. We will provide a more detailed introduction to these datasets in Appendix.

\subsubsection{Evaluation Criteria}

The evaluation metrics employed in our study are the widely utilized Pearson Linear Correlation Coefficient (PLCC) and Spearman's Rank Correlation Coefficient (SRCC), both of which range from 0 to 1. Values approaching 1 denote a higher degree of prediction relevance.

\subsubsection{Experimental Details}

Our experimental methodology closely follows the training strategy outlined in DEIQT\cite{DEIQT}, where input images are randomly cropped into ten patches, each with a resolution of 224×224. We employed three variants of the VMamba architecture: QMamba-B, QMamba-S, and QMamba-T. Both QMamba-B and QMamba-S feature an encoder with a depth of 15 blocks, with QMamba-B incorporating an embedding dimension of 128, and QMamba-S using an embedding dimension of 96. In contrast, QMamba-T is designed with a reduced depth of 4 blocks and an embedding dimension of 96. Training procedures leveraged weights pre-trained on the ImageNet-1K dataset, spanning a total of 9 epochs. Batch sizes were adjusted according to the respective dataset sizes, \egno, 32 for LIVEC and 128 for KonIQ. We used the AdamW optimizer for training, with the learning rate set to $2 \times 10^{-4}$ and a decay factor of 10 applied every 3 epochs. We compared the performance of ResNet\cite{ResNet}, ViT\cite{ViT}, and Swin Transformer\cite{SwinTransformer}, all of which were implemented using the official versions and loaded with pre-trained weights. To ensure fairness, other experimental settings were kept as consistent as possible. All experiments were conducted using multiple NVIDIA RTX 4090 GPUs.

\subsection{A Comparison Between Different IQA Backbones}

\subsubsection{Task-specific IQA}

 We conducted comprehensive training and testing across the ten datasets previously introduced, drawing analytical comparisons based on results reported by existing methods. For this task, $80\%$ of the images in each dataset were used for training, while the remaining $20\%$ were reserved for testing. Given the predominance of BIQA methods targeting synthetic and authentic datasets, Table \ref{table: task_specfic_result} presents a detailed comparative analysis of state-of-the-art (SOTA) methods and popular architectures such as ResNet, ViT, and Swin Transformer, in comparison to our QMamba architecture. This comparison elucidates performance discrepancies and computational complexities across different parameter configurations. The results in the table demonstrate that the LQMamba configuration achieves optimal performance across six datasets. Compared to existing IQA methods with similar parameter counts, QMamba-T exhibits lower GFLOPS and superior performance, confirming the efficiency and reduced computational complexity of the Mamba architecture. Comparative results for the two AIGC datasets are documented in the appendix.

\begin{table}[h!]
\centering
\begin{threeparttable}

\renewcommand{\arraystretch}{1.3}
\resizebox{\linewidth}{!}{
\begin{tabular}{c|c c c c}
\Xhline{1.2pt}
\multicolumn{5}{c}{Mixed Training} \\ \Xhline{1pt}
\multicolumn{5}{c}{LIVE \& KADID \& LIVEC \& KonIQ \& AGIQA3K \& AIGCIQA2023} \\ \Xhline{1pt}
\multicolumn{1}{c|}{Method} & \multicolumn{1}{c}{Parameters} & \multicolumn{1}{c}{GFLOPS} & \multicolumn{1}{c}{PLCC\_Average} & \multicolumn{1}{c}{SRCC\_Average} \\ \Xhline{1pt}
\multicolumn{1}{c|}{ResNet-50}  & 23.51M & 4.11G  & 0.878  & 0.853         \\
\multicolumn{1}{c|}{ViT-S}      & 21.67M & 4.61G  & 0.891 &  0.867        \\
\multicolumn{1}{c|}{Swin-T}     & 27.52M & 4.51G  & 0.900 &    0.883      \\
\Xhline{1pt}
\multicolumn{1}{c|}{DEIQT}      & 24.04M & 5.41G  & 0.895 &   0.873       \\ 
\multicolumn{1}{c|}{LoDa}       & 8.93M* & 23.68G & 0.876 &  0.855        \\
\Xhline{1pt}
\multicolumn{1}{c|}{QMamba-T}   & 27.99M & 4.47G  & 0.905 &  0.886        \\
\multicolumn{1}{c|}{LQMamba-T}  & 29.87M & 4.44G  & \textbf{0.909} &  \textbf{0.888}      \\
\Xhline{1.2pt}
\end{tabular}
}
\begin{tablenotes}
\item[\tiny*Trainable parameters.]
\end{tablenotes}
\end{threeparttable}
\caption{Performance Comparison for Universal IQA.}
\label{table: Universal Results}
\end{table}

\begin{table*}[h!]
\centering
\renewcommand{\arraystretch}{1.2} 
\resizebox{\linewidth}{!}{
\setlength{\tabcolsep}{0.7mm}{
\begin{tabular}{ccccccccc|cccccccc|c}
\Xhline{1.2pt}
Train                                                                                                                & \multicolumn{8}{c|}{KonIQ \& LIVEC (Authentic)}                                                                                & \multicolumn{8}{c|}{KADID \& LIVE (Synthetic)}                                                                                          \\ \Xhline{1pt}
Test                                                                                                                 & \multicolumn{2}{c}{KADID} & \multicolumn{2}{c}{LIVE} & \multicolumn{2}{c}{AIGC2023} & \multicolumn{2}{c|}{AGIQA3K} & \multicolumn{2}{c}{KonIQ} & \multicolumn{2}{c}{LIVEC} & \multicolumn{2}{c}{AIGC2023} & \multicolumn{2}{c|}{AGIQA3K}         \\ \Xhline{1pt}
\multicolumn{1}{c|}{Fine-tuning Method}                                                                              & PLCC        & SRCC        & PLCC        & SRCC       & PLCC          & SRCC         & PLCC          & SRCC         & PLCC        & SRCC        & PLCC        & SRCC        & PLCC          & SRCC         & PLCC          & SRCC         & Average        \\ \Xhline{1pt}
\multicolumn{1}{c|}{DEIQT}                                                                         & 0.583       & 0.595       & 0.558       & 0.564      & 0.802         & 0.794        & 0.698         & 0.632        & 0.518       & 0.526       & 0.579       & 0.532       & 0.580         & 0.575   & 0.512      & \multicolumn{1}{c|}{0.506} & 0.601  \\

\multicolumn{1}{c|}{LoDa}                                                                         & 0.600       & 0.600       & 0.733       & 0.729      & 0.808         & 0.800        & 0.715         & 0.650        & 0.527       & 0.513       & 0.558       & 0.499       & 0.592         & 0.586 & 0.564         & \multicolumn{1}{c|}{0.541} & 0.622 \\
\multicolumn{1}{c|}{Without\_tuning}                                                                         & 0.535       & 0.501       & 0.726       & 0.744      & 0.789         & 0.784        & 0.726         & 0.744        & 0.553       & 0.549       & 0.595       & 0.550       & 0.600         & 0.612        & 0.675         & 0.649        & 0.642          \\
\multicolumn{1}{c|}{Lin\_Probe(R)}                                                                                       & 0.606       & 0.581       & 0.783       & 0.811      & 0.824         & 0.804        & 0.769         & 0.695        & 0.871       & 0.848       & 0.785       & 0.751       & 0.803         & 0.788        & 0.771         & 0.723        & 0.755          \\ 
\multicolumn{1}{c|}{ \begin{tabular}[c]{@{}c@{}}Full\_tuning \textbf{(93.79M)} \end{tabular}} & 0.936       & 0.930       & 0.941       & 0.940      & 0.885         & 0.863        & 0.910         & 0.851        & 0.943       & 0.928       & 0.899       & 0.873       & 0.880         & 0.859        & 0.910         & 0.856        & \textbf{0.908} \\
\multicolumn{1}{c|}{\begin{tabular}[c]{@{}c@{}}StylePrompt \textbf{(3.83M)}\end{tabular}}                                   & 0.920       & 0.912       & 0.949       & 0.948      & 0.877         & 0.854        & 0.908         & 0.854        & 0.932       & 0.913       & 0.888       & 0.866       & 0.880         & 0.865        & 0.906         & 0.852        & \textbf{0.901} \\
\multicolumn{1}{c|}{StylePrompt \& R}                                                                                & 0.921       & 0.913       & 0.944       & 0.945      & 0.877         & 0.853        & 0.906         & 0.847        & 0.931       & 0.912       & 0.889       & 0.860       & 0.876         & 0.857        & 0.905         & 0.849        & 0.898          \\ \Xhline{1.2pt}
\end{tabular}
}
}
\caption{Performance Comparison for Transferable IQA.}
\label{table: Transferable Results}
\end{table*}

\subsubsection{Universal IQA}

We evaluated the effectiveness of the QMamba architecture for universal tasks by employing mixed training across six different datasets: two synthetic datasets (LIVE, KADID), two authentic datasets (LIVEC, KonIQ), and two AIGC datasets (AIGC2023, AGIQA3k). For each dataset, $20\%$ of the data was reserved for performance testing. The average results of models with similar scales are presented in Table \ref{table: Universal Results}, with detailed results provided in Appendix. These findings highlight strong multi-tasking capabilities of Mamba. Compared to several mainstream models, QMamba performed well across most datasets, confirming its effectiveness in handling general tasks.

\subsubsection{Analysis}
Our investigation into the efficacy of QMamba for IQA tasks reveals two key insights through analysis of the KADID dataset (7 distortion types) and cross-dataset validation. As shown in Figure \ref{fig: t-SNE}, t-SNE visualization demonstrates superior distortion discrimination of QMamba: 1) Tightly clustered features for each distortion type, 2) Clear separation between dissimilar artifacts, and 3) Minimal inter-class overlap compared to ViT's partial merging and CNN/Swin architectures' significant feature entanglement.

This discriminative capability directly impacts practical performance. While QMamba demonstrates modest gains on simpler datasets such as LIVE and CSIQ, which contain only 4 to 5 distortion types, it achieves substantial improvements on more complex benchmarks like TID2013 and KADID, which include 24 to 25 distortion types. The architecture based on state-space modeling enables adaptive frequency processing through a selective scanning mechanism, dynamically emphasizing distortion-critical patterns while suppressing irrelevant features. This approach stands in contrast to convolutional networks with fixed receptive fields and Transformers that tend to over-mix local characteristics through global attention.

Our findings indicate that QMamba is well-suited for new quality assessment scenarios involving complex and mixed distortions, such as those introduced by neural compression or generative models. In these cases, traditional architectures often fail to capture subtle or entangled artifacts, while the selective modeling in QMamba provides better adaptability and robustness.

\subsection{Efficient Transfer Learning for Mamba-Based IQA}

In the context of transferable IQA tasks, we conducted domain-specific training using synthetic datasets (LIVE, KADID), authentic datasets (LIVEC, KonIQ), and AIGC datasets (AIGC2023, AGIQA3K). After training in one domain, models were directly transferred and tested on datasets from the other two domains. We employed the StylePrompt technique, as illustrated in Figure~\ref{fig: QMamba and StylePrompt}, where the architecture was kept intact by freezing all model parameters and fine-tuning only the StylePrompt module, which involved approximately $4\%$ of the total parameter count. This approach achieved performance levels comparable to those obtained through full-parameter training. The outcomes, presented in Table~\ref{table: Transferable Results}, clearly demonstrate the effectiveness and efficiency of the proposed StylePrompt method for transferable IQA tasks. The value of "Average" represents the mean performance across all domain transfers, with more detailed results provided in the Appendix.

\subsection{Ablation Study}

\subsubsection{Different Scanning Method}

As shown in the task-specific results (Table~\ref{table: task_specfic_result}) and universal evaluation results (Table~\ref{table: Universal Results}), models adopting local scanning consistently outperform those based on cross-scanning strategies. This motivates our exploration of local scanning mechanisms, as employed in LQMamba.

Although the average performance gap between QMamba and LQMamba may appear small, further analysis reveals that LQMamba performs better on most individual datasets. The marginal overall gain is primarily due to relatively lower improvements on simpler datasets such as LIVE and CSIQ, which contain fewer distortion types and less diverse content. In contrast, on more challenging datasets like TID2013 and KADID-10k, which feature a wide variety of fine-grained distortions, LQMamba shows clear advantages (\egno, SRCC on TID2013: 0.964 vs. 0.949; on KADID: 0.941 vs. 0.932). These results highlight the effectiveness of the local scanning design in complex, distortion-rich scenarios where precise modeling of local artifacts is critical.

\subsubsection{Different Model Scale}

In our empirical analysis, as documented in Table \ref{table: task_specfic_result}, we discern that while QMamba-Base exhibits exceptionally robust quality perception capabilities, QMamba-Small either matches or exceeds the performance of QMamba-Base across the majority of the datasets. Although QMamba-Tiny displays a modest decline in performance metrics, it still delivers results that are competitive with current SOTA methods, solely utilizing the capabilities of QMamba-Tiny.

\begin{table}[h!]
\centering
\small
\renewcommand{\arraystretch}{1.2}
\setlength{\tabcolsep}{0.5mm}{
\begin{tabular}{>{\centering\arraybackslash}p{3cm}|c|c|c}
\Xhline{1.0pt}
Tuning Strategy         & Parameters & PLCC\_Avg.   & SRCC\_Avg.  \\ \Xhline{1pt}
SSF                     & 6.1M       & 0.750       & 0.735      \\
Crossattn\_Prompt       & 12.17M     & 0.806       & 0.772      \\
Conv\_Prompt            & 28.33M     & 0.883       & 0.856      \\
StylePrompt\textbf{(ours)} & 3.83M    & \textbf{0.911} & \textbf{0.890} \\ \Xhline{1.0pt}
\end{tabular}
}
\caption{Ablation Study on Different Prompt Tuning Strategies}
\label{table: Ablation Prompt Tuning Strategies}
\end{table}

\subsubsection{Different Prompt Tuning Strategies}
To validate the effectiveness of the StylePrompt Generation (SPG) process, we conducted tests by directly learning a set of affine parameters \( \gamma \) and \( \beta \) to modulate the original features, rather than using prompts for learning, similar to the SSF\cite{SSF}. Additionally, to assess the effectiveness of the StylePrompt interaction method, we explored various interaction strategies during tuning, including convolutional prompt interaction and cross-attention prompt interaction. The results in Table \ref{table: Ablation Prompt Tuning Strategies} demonstrate the superior efficiency and performance of StylePrompt.

In addition, we conducted several ablation studies related to the design of the prompts, including variations in the number and shape of the prompts. The detailed results of these experiments are provided in Appendix.

\subsection{Discussion and Future Work}

Looking ahead, the sequential modeling nature and efficiency of our SSM-based architecture make it a promising candidate for extension to video and audio quality assessment. Given the higher temporal complexity in VQA and the inherent sequential structure of audio signals, our method offers a unified and lightweight foundation for future research across visual and multimodal quality evaluation tasks.

\section{Conclusion}

In this paper, we introduced QMamba, a novel state space framework for image quality assessment that integrates task-specific evaluation, universal perception, and cross-domain transferability. Extensive experiments demonstrate that QMamba consistently outperforms established vision models, achieving significant improvements in accuracy while requiring substantially lower computational resources. The proposed StylePrompt mechanism enables robust cross-domain adaptation through lightweight dynamic feature recalibration, setting new benchmarks for IQA and demonstrating strong practical potential in real-world applications.

\newpage

\nocite{langley00}

\bibliography{Qmamba}
\bibliographystyle{icml2025}

\newpage
\appendix

\section{Appendix / supplemental material}

\subsection{Detailed Description of the Datasets}
We evaluate the performance of the proposed method across eight widely recognized BIQA datasets, comprising both synthetic and authentic datasets. The synthetic datasets include LIVE\cite{live2}, CSIQ\cite{CSIQ}, TID2013\cite{Tid2013}, and KADID-10k\cite{Kadid}. These datasets feature a small number of pristine images that are synthetically distorted using various techniques such as JPEG compression and Gaussian blurring. Specifically, LIVE contains 799 images with 5 types of distortion, CSIQ includes 866 images with 6 distortion types, TID2013 comprises 3000 images with 24 distortion types, and KADID-10k includes 10125 images with 25 distortion types.

On the other hand, the authentic datasets include LIVEC\cite{livec}, KonIQ-10k\cite{Koniq}, SPAQ\cite{SPAQ}, and FLIVE\cite{LIVEFB}. LIVEC consists of 1,162 images with diverse authentic distortions captured by mobile devices. KonIQ-10k is composed of 10,073 images selected from YFCC-100M, covering a wide range of distortions such as brightness, colorfulness, contrast, noise, and sharpness. SPAQ contains 11,125 images collected by various smartphones, representing a large variety of scene categories. FLIVE, the largest in-the-wild IQA dataset to date, contains 39,810 real-world images with diverse content, sizes, and aspect ratios.

In response to the rapid development of AI-generated content, we also employed two additional datasets: AIGCIQA2023\cite{AIGCIQA2023} and AGIQA3K\cite{AGIQA3K}. AIGCIQA2023 contains over 2000 images generated by six state-of-the-art text-to-image models, evaluated through a subjective experiment on quality, authenticity, and correspondence. AGIQA3K consists of 2,982 images from GAN, autoregression, and diffusion-based models, with annotations for perceptual quality and text-to-image alignment. 

\subsection{Additional Experimental Results}
\subsubsection{Task-specific IQA results on AIGC datasets}
We present the test results of various backbones on the AIGC datasets in Table \ref{table: AIGC_task_specific_detail}, with bold indicating the best results.

\begin{table}[h!]
\centering
\renewcommand{\arraystretch}{1.5}
\resizebox{\linewidth}{!}{
\begin{tabular}{ccccccccc}
\Xhline{2pt}
Model                                   &                             & \multicolumn{2}{c}{AGIQA3K}     & \multicolumn{2}{c}{AIGCIQA2023} &               \\ \Xhline{1pt}
\multicolumn{1}{c|}{}                   & \multicolumn{1}{c|}{GFLOPS} & PLCC           & SRCC           & PLCC           & SRCC           & Average        \\ \Xhline{1pt}
\multicolumn{1}{c|}{ResNet-50(23.51M)}  & \multicolumn{1}{c|}{4.11G}  & 0.901          & 0.840          & 0.795          & 0.797          & 0.833          \\
\multicolumn{1}{c|}{ResNet-101(42.50M)} & \multicolumn{1}{c|}{7.83G}  & 0.907          & 0.847          & 0.834          & 0.831          & 0.855          \\
\multicolumn{1}{c|}{ResNet-152(58.15M)} & \multicolumn{1}{c|}{11.53G} & 0.901          & 0.832          & 0.841          & 0.834          & 0.852          \\ \hline
\multicolumn{1}{c|}{ViT-T(5.52M)}       & \multicolumn{1}{c|}{1.26G}  & 0.865          & 0.787          & 0.760          & 0.766          & 0.795          \\
\multicolumn{1}{c|}{ViT-S(21.67M)}      & \multicolumn{1}{c|}{4.61G}  & 0.891          & 0.819          & 0.842          & 0.822          & 0.844          \\
\multicolumn{1}{c|}{ViT-B(85.80M)}      & \multicolumn{1}{c|}{17.58G} & 0.897          & 0.830          & 0.853          & 0.835          & 0.854          \\ \hline
\multicolumn{1}{c|}{Swin-T(27.52M)}     & \multicolumn{1}{c|}{4.51G}  & 0.906          & 0.847          & 0.867          & 0.844          & 0.866          \\
\multicolumn{1}{c|}{Swin-S(48.84M)}     & \multicolumn{1}{c|}{8.77G}  & 0.908          & 0.849          & 0.875          & 0.857          & 0.872          \\
\multicolumn{1}{c|}{Swin-B(86.74M)}     & \multicolumn{1}{c|}{15.47G} & 0.909          & 0.852          & 0.886          & 0.863          & 0.878          \\ \hline
\multicolumn{1}{c|}{QMamba-T (27.99M)}  & \multicolumn{1}{c|}{4.47G}  & 0.913          & 0.858          & 0.888          & 0.873          & 0.883          \\
\multicolumn{1}{c|}{QMamba-S (49.37M)}  & \multicolumn{1}{c|}{8.71G}  & 0.912          & 0.858          & \textbf{0.889} & \textbf{0.875} & \textbf{0.884} \\
\multicolumn{1}{c|}{QMamba-B (87.53M)}  & \multicolumn{1}{c|}{15.35G} & 0.914          & 0.861          & 0.886          & 0.868          & 0.882          \\ \hline
\multicolumn{1}{c|}{LQMamba-T(29.87M)}  & \multicolumn{1}{c|}{4.44G}  & 0.914          & 0.862          & 0.884          & 0.868          & 0.882          \\
\multicolumn{1}{c|}{LQMamba-S(52.91M)}  & \multicolumn{1}{c|}{8.66G}  & 0.913          & \textbf{0.864} & 0.888          & 0.869          & \textbf{0.884} \\
\multicolumn{1}{c|}{LQMamba-B(93.79M)}  & \multicolumn{1}{c|}{15.30G} & \textbf{0.915} & 0.858          & 0.888          & 0.871          & 0.883          \\ \Xhline{2pt}
\end{tabular}
}
\caption{Results of AIGC for task-specific IQA on various backbones}
\label{table: AIGC_task_specific_detail}
\end{table}

\subsubsection{Universal IQA detailed data}
We provide the detailed results of Universal IQA in Table~\ref{table: Universal Results detail}.

\begin{table*}[h!]
\centering
\renewcommand{\arraystretch}{1.3}
\resizebox{\linewidth}{!}{
\begin{tabular}{cclllllllllllll}
\Xhline{1.5pt}
Train                           & \multicolumn{1}{l}{}        & \multicolumn{12}{c}{LIVE \& KADID \& LIVEC \& KonIQ \& AGIQA3K \& AIGCIQA2023}                                                                                                                                                                                                                                                                            &                \\ \Xhline{1pt}
Test                            & \multicolumn{1}{l}{}        & \multicolumn{2}{c}{LIVE}                            & \multicolumn{2}{c|}{KADID}                                     & \multicolumn{2}{c}{LIVEC}                           & \multicolumn{2}{c|}{KonIQ}                                     & \multicolumn{2}{c}{AGIQA3K}                           & \multicolumn{2}{c}{AIGCIQA2023}                     &                \\ \Xhline{1pt}
\multicolumn{1}{c|}{}           & \multicolumn{1}{l|}{GFLOPS} & \multicolumn{1}{c}{PLCC} & \multicolumn{1}{c}{SRCC} & \multicolumn{1}{c}{PLCC} & \multicolumn{1}{c|}{SRCC}           & \multicolumn{1}{c}{PLCC} & \multicolumn{1}{c}{SRCC} & \multicolumn{1}{c}{PLCC} & \multicolumn{1}{c|}{SRCC}           & \multicolumn{1}{c}{PLCC} & \multicolumn{1}{c}{SRCC} & \multicolumn{1}{c}{PLCC} & \multicolumn{1}{c}{SRCC} & Average        \\ \Xhline{1pt}
\multicolumn{1}{c|}{ResNet-50}  & \multicolumn{1}{c|}{4.11G}  & 0.916                    & 0.908                    & 0.868                    & \multicolumn{1}{l|}{0.869}          & 0.885                    & 0.841                    & 0.896                    & \multicolumn{1}{l|}{0.869}          & 0.884                    & 0.813                    & 0.821                    & 0.815                    & 0.865          \\

\multicolumn{1}{c|}{ViT-S}      & \multicolumn{1}{c|}{4.61G}  & 0.921                    & 0.918                    & 0.920                    & \multicolumn{1}{l|}{0.917}          & 0.883                    & 0.828                    & 0.911                    & \multicolumn{1}{l|}{0.892}          & 0.885                    & 0.822                    & 0.825                    & 0.824                    & 0.879          \\

\multicolumn{1}{c|}{Swin-T}     & \multicolumn{1}{c|}{4.51G}  & 0.927                    & \textbf{0.927}                    & 0.925                    & \multicolumn{1}{l|}{0.921}          & 0.889                    & 0.863                    & 0.929                    & \multicolumn{1}{l|}{0.918}          & 0.897                    & 0.840                    & 0.834                    & 0.828                    & 0.892          \\
\multicolumn{1}{c|}{DEIQT} & \multicolumn{1}{c|}{4.68G} & 0.907 & 0.906 & 0.898 & \multicolumn{1}{l|}{0.895} & 0.896 & 0.855 & 0.928 & \multicolumn{1}{l|}{0.906} & 0.902 & 0.841 & 0.839 & \textbf{0.835}  & 0.884 \\
\multicolumn{1}{c|}{LoDa} & \multicolumn{1}{c|}{23.68G} & 0.895 & 0.902 & 0.907 & \multicolumn{1}{l|}{0.900} & 0.872 & 0.842 & 0.900 & \multicolumn{1}{l|}{0.875} & 0.874 & 0.809 & 0.809 & 0.801  & 0.866 \\

\multicolumn{1}{c|}{QMamba-T}   & \multicolumn{1}{c|}{4.47G}  & 0.923                    & 0.923                    & 0.938                    & \multicolumn{1}{l|}{0.932}          & 0.898                    & 0.863                    & 0.932                    & \multicolumn{1}{l|}{0.917}          & 0.906                    & 0.851                    & 0.835                    & 0.830                    & 0.896          \\

\multicolumn{1}{c|}{LQMamba-T}  & \multicolumn{1}{c|}{4.44G}  & \textbf{0.929}                    & 0.926                    & \textbf{0.943}                    & \multicolumn{1}{l|}{\textbf{0.939}}          & \textbf{0.899}                    & \textbf{0.863}                    & \textbf{0.936}                    & \multicolumn{1}{l|}{\textbf{0.921}}          & \textbf{0.908}                    & \textbf{0.853}                    & \textbf{0.840}                    & 0.828                    & \textbf{0.899}          \\
\Xhline{1.5pt}
\end{tabular}
}
\caption{Performance comparison for universal IQA.}
\label{table: Universal Results detail}

\end{table*}

\subsubsection{Transferable IQA results trained on the AIGC domain}

We provide the detailed results of Transferable IQA trained in the AIGC domain in Table~\ref{table: Transferable Results_detail}.

\begin{table*}[h!]
\centering
\scriptsize 
\renewcommand{\arraystretch}{1.2} 
\resizebox{\linewidth}{!}{
\setlength{\tabcolsep}{0.7mm}{
\begin{tabular}{cccccccccc}
\hline 
Train                                                                                                                & \multicolumn{8}{c}{AIGC2023 \& AGIQA3K (AIGC)}                                                                                                                 \\ \hline
Test                                                                                                                 & \multicolumn{2}{c}{KonIQ} & \multicolumn{2}{c}{LIVEC} & \multicolumn{2}{c}{KADID} & \multicolumn{2}{c}{LIVE}                                          \\ \hline
\multicolumn{1}{c|}{Fine-tuning Method}                                                                              & PLCC        & SRCC        & PLCC        & SRCC        & PLCC        & SRCC        & PLCC        & \multicolumn{1}{c|}{SRCC}                          & Average        \\ \hline
\multicolumn{1}{c|}{DEIQT} & 0.578 & 0.505 & 0.684 & 0.638 & 0.498 & 0.464 & 0.752 & \multicolumn{1}{c|}{0.750}  &0.601\\
\multicolumn{1}{c|}{LoDa} & 0.564 & 0.500 & 0.542 & 0.548 & 0.555 & 0.524 & 0.840 & \multicolumn{1}{c|}{0.833}  &0.622\\

\multicolumn{1}{c|}{Without\_tuning}                                                                         & 0.636       & 0.587       & 0.667       & 0.637       & 0.460       & 0.425       & 0.815 & \multicolumn{1}{c|}{0.838}                         & 0.642          \\
\multicolumn{1}{c|}{Lin\_Probe(R)}                                                                                       & 0.818       & 0.792       & 0.790       & 0.761       & 0.570       & 0.540       & 0.799 & \multicolumn{1}{c|}{0.826}                         & 0.755          \\ 
\multicolumn{1}{c|}{ \begin{tabular}[c]{@{}c@{}}Full\_tuning \textbf{(93.79M)} \end{tabular}} & 0.943       & 0.927       & 0.910       & 0.878       & 0.937       & 0.932       & 0.938 & \multicolumn{1}{c|}{0.933} & \textbf{0.908} \\
\multicolumn{1}{c|}{\begin{tabular}[c]{@{}c@{}}StylePrompt \textbf{(3.83M)}\end{tabular}}                                   & 0.929       & 0.909       & 0.902       & 0.871       & 0.918       & 0.913       & 0.926 & \multicolumn{1}{c|}{0.922}                         & \textbf{0.901} \\
\multicolumn{1}{c|}{StylePrompt \& R}                                                                                & 0.929       & 0.909       & 0.899       & 0.859       & 0.918       & 0.911       & 0.926 & \multicolumn{1}{c|}{0.927}                         & 0.898          \\ \hline
\end{tabular}
}
}
\caption{Performance comparison for transferable IQA (trained on AIGC).}
\label{table: Transferable Results_detail}
\end{table*}

\subsubsection{Detailed results of ablation studies}

\begin{itemize}
    \item \textbf{Ablation study on different prompt tuning strategies.}\\
    We have provided the detailed experimental data for the ablation study on different prompt tuning strategies in Table \ref{table: Ablation Prompt Tuning Strategies_detail}.
    \item \textbf{Ablation study results for different prompt designs.}\\
    We investigated the impact of the number of prompt components on performance and found that setting the number to six yields optimal results. Additionally, we explored whether varying the spatial dimensions and sizes of prompts would enhance performance. Our findings show that a spatial size of (1,1), focusing solely on the channel dimension, offers the best results. The outcomes of these ablation studies across multiple datasets are presented in Tables~\ref{table: Ablation N} and~\ref{table: Ablation Size}.
\end{itemize}

\begin{table*}[h!]
\centering
\renewcommand{\arraystretch}{2}
\resizebox{\linewidth}{!}{
\setlength{\tabcolsep}{0.5mm}{
\large
\begin{tabular}{ccccccccc|cccccccc|ccccccccc}

\Xhline{1.5pt}
Train                                          & \multicolumn{8}{c|}{KonIQ \& LIVEC}                                                                                                   & \multicolumn{8}{c|}{KADID \& LIVE}                                                                                                    & \multicolumn{8}{c}{AIGC2023 \& AGIQA3K}                                                                                                                &                \\ \Xhline{1pt}
Test                                           & \multicolumn{2}{c}{KADID}       & \multicolumn{2}{c}{LIVE}        & \multicolumn{2}{c}{AIGC2023}    & \multicolumn{2}{c|}{AGIQA3K}    & \multicolumn{2}{c}{KonIQ}       & \multicolumn{2}{c}{LIVEC}       & \multicolumn{2}{c}{AIGC2023}    & \multicolumn{2}{c|}{AGIQA3K}    & \multicolumn{2}{c}{KonIQ}       & \multicolumn{2}{c}{LIVEC}       & \multicolumn{2}{c}{KADID}       & \multicolumn{2}{c}{LIVE}                             &                \\ \Xhline{1pt}
\multicolumn{1}{c|}{Style and Prompt method}   & PLCC           & SRCC           & PLCC           & SRCC           & PLCC           & SRCC           & PLCC           & SRCC           & PLCC           & SRCC           & PLCC           & SRCC           & PLCC           & SRCC           & PLCC           & SRCC           & PLCC           & SRCC           & PLCC           & SRCC           & PLCC           & SRCC           & PLCC           & \multicolumn{1}{c|}{SRCC}           & Average        \\ \Xhline{1pt}
\multicolumn{1}{c|}{SSF\textbf{(6.1M)}}                 & 0.718          & 0.700          & 0.735          & 0.769          & 0.804          & 0.801          & 0.714          & 0.667          & 0.863          & 0.829          & 0.643          & 0.598          & 0.668          & 0.675          & 0.714          & 0.685          & 0.866          & 0.828          & 0.730          & 0.706          & 0.724          & 0.708          & 0.820          & \multicolumn{1}{c|}{0.856}          & 0.743          \\
\multicolumn{1}{c|}{Crossattn\_Prompt\textbf{(12.17M)}} & 0.843          & 0.830          & 0.898          & 0.897          & 0.783          & 0.752          & 0.821          & 0.706          & 0.843          & 0.816          & 0.641          & 0.612          & 0.802          & 0.782          & 0.833          & 0.726          & 0.842          & 0.818          & 0.644          & 0.616          & 0.831          & 0.815          & 0.894          & \multicolumn{1}{c|}{0.897}          & 0.789          \\
\multicolumn{1}{c|}{Conv\_Prompt\textbf{(28.33M)}}      & 0.911          & 0.910          & 0.945          & 0.946          & \textbf{0.889} & \textbf{0.820} & 0.890          & 0.825          & 0.897          & 0.881          & 0.797          & 0.761          & 0.864          & 0.827          & 0.861          & 0.828          & 0.898          & 0.884          & 0.805          & 0.756          & 0.901          & 0.899          & \textbf{0.932} & \multicolumn{1}{c|}{\textbf{0.933}} & 0.869 \\
\multicolumn{1}{c|}{StylePrompt\textbf{(ours)}\textbf{(3.83M)}}          & \textbf{0.920} & \textbf{0.912} & \textbf{0.949} & \textbf{0.948} & 0.877          & 0.854          & \textbf{0.908} & \textbf{0.854} & \textbf{0.932} & \textbf{0.913} & \textbf{0.888} & \textbf{0.866} & \textbf{0.880} & \textbf{0.865} & \textbf{0.906} & \textbf{0.852} & \textbf{0.929} & \textbf{0.909} & \textbf{0.902} & \textbf{0.871} & \textbf{0.918} & \textbf{0.913} & 0.926          & \multicolumn{1}{c|}{0.922}          & \textbf{0.901} \\ \Xhline{1.5pt}
\end{tabular}
}
}
\caption{Ablation study on different prompt tuning strategies}

\label{table: Ablation Prompt Tuning Strategies_detail}
\end{table*}

\begin{table*}[h!]
\centering
\renewcommand{\arraystretch}{2}
\resizebox{\linewidth}{!}{
\setlength{\tabcolsep}{0.5mm}{
\large
\begin{tabular}{ccccccccc|cccccccc|ccccccccc}
\Xhline{2pt}
Train                     & \multicolumn{8}{c|}{KonIQ \& LIVEC}                                                                                                   & \multicolumn{8}{c|}{KADID \& LIVE}                                                                                                    & \multicolumn{8}{c}{AIGC2023 \& AGIQA3K}                                                                                                                    &                 \\ \Xhline{1pt}
Test                      & \multicolumn{2}{c}{KADID}       & \multicolumn{2}{c}{LIVE}        & \multicolumn{2}{c}{AIGC2023}    & \multicolumn{2}{c|}{AGIQA3K}    & \multicolumn{2}{c}{KonIQ}       & \multicolumn{2}{c}{LIVEC}       & \multicolumn{2}{c}{AIGC2023}    & \multicolumn{2}{c|}{AGIQA3K}    & \multicolumn{2}{c}{KonIQ}       & \multicolumn{2}{c}{LIVEC}       & \multicolumn{2}{c}{KADID}       & \multicolumn{2}{c}{LIVE}                             &                 \\ \Xhline{1pt}
\multicolumn{1}{c|}{}     & PLCC           & SRCC           & PLCC           & SRCC           & PLCC           & SRCC           & PLCC           & SRCC           & PLCC           & SRCC           & PLCC           & SRCC           & PLCC           & SRCC           & PLCC           & SRCC           & PLCC           & SRCC           & PLCC           & SRCC           & PLCC           & SRCC           & PLCC           & \multicolumn{1}{c|}{SRCC}           & Average         \\ \Xhline{1pt}
\multicolumn{1}{c|}{N=1}  & 0.914          & 0.907          & 0.943          & 0.944          & 0.877          & 0.850          & 0.900          & 0.833          & 0.924          & 0.906          & 0.873          & 0.846          & 0.872          & 0.850          & 0.898          & 0.834          & 0.924          & 0.907          & 0.890          & 0.855          & 0.904          & 0.899          & \textbf{0.940} & \multicolumn{1}{c|}{\textbf{0.937}} & 0.8928          \\
\multicolumn{1}{c|}{N=2}  & 0.913          & 0.905          & 0.938          & 0.939          & 0.874          & 0.851          & 0.895          & 0.837          & 0.930          & 0.912          & 0.886          & 0.853          & 0.871          & 0.851          & 0.892          & 0.835          & 0.927          & 0.905          & 0.902          & 0.865          & 0.911          & 0.906          & 0.926          & \multicolumn{1}{c|}{0.922}          & 0.8936          \\
\multicolumn{1}{c|}{N=4}  & 0.920          & 0.913          & 0.944          & 0.949          & 0.875          & 0.851          & 0.904          & 0.840          & 0.929          & 0.913          & 0.877          & 0.849          & 0.876          & 0.857          & 0.904          & 0.842          & 0.927          & 0.904          & 0.885          & 0.847          & 0.913          & 0.907          & 0.925          & \multicolumn{1}{c|}{0.931}          & 0.8951          \\
\multicolumn{1}{c|}{N=6}  & 0.920          & 0.912          & \textbf{0.949} & \textbf{0.948} & 0.877          & 0.854          & \textbf{0.908} & \textbf{0.854} & \textbf{0.932} & \textbf{0.913} & \textbf{0.888} & \textbf{0.866} & \textbf{0.880} & \textbf{0.865} & \textbf{0.906} & \textbf{0.852} & \textbf{0.929} & \textbf{0.909} & \textbf{0.902} & \textbf{0.871} & \textbf{0.918} & \textbf{0.913} & 0.926          & \multicolumn{1}{c|}{0.922}          & \textbf{0.9006} \\
\multicolumn{1}{c|}{N=8}  & \textbf{0.920} & \textbf{0.914} & 0.930          & 0.934          & 0.874          & 0.851          & 0.903          & 0.841          & 0.930          & 0.912          & 0.873          & 0.843          & 0.876          & 0.854          & 0.899          & 0.841          & 0.927          & 0.906          & 0.889          & 0.859          & 0.918          & 0.912          & 0.922          & \multicolumn{1}{c|}{0.926}          & 0.8939          \\
\multicolumn{1}{c|}{N=10} & 0.919          & 0.913          & 0.930          & 0.936          & \textbf{0.877} & \textbf{0.856} & 0.901          & 0.840          & 0.929          & 0.910          & 0.880          & 0.849          & 0.877          & 0.853          & 0.900          & 0.840          & 0.925          & 0.906          & 0.889          & 0.851          & 0.918          & 0.914          & 0.923          & \multicolumn{1}{c|}{0.924}          & 0.8942          \\ \Xhline{2pt}
\end{tabular}
}
}
\caption{Ablation study results for different numbers of prompts (N).}

\label{table: Ablation N}

\end{table*}

\begin{table*}[h!]
\centering
\renewcommand{\arraystretch}{2}
\resizebox{\linewidth}{!}{
\setlength{\tabcolsep}{0.5mm}{
\large
\begin{tabular}{ccccccccc|cccccccc|ccccccccc}
\Xhline{2pt}
Train                          & \multicolumn{8}{c|}{KonIQ \& LIVEC}                                                                                           & \multicolumn{8}{c|}{KADID \& LIVE}                                                                                            & \multicolumn{8}{c}{AIGC2023 \& AGIQA3K}                                                                                                            &                \\ \Xhline{1pt}
Test                           & \multicolumn{2}{c}{KADID}       & \multicolumn{2}{c}{LIVE}        & \multicolumn{2}{c}{AIGC2023}    & \multicolumn{2}{c|}{AGIQA3K}    & \multicolumn{2}{c}{KonIQ}       & \multicolumn{2}{c}{LIVEC}       & \multicolumn{2}{c}{AIGC2023}    & \multicolumn{2}{c|}{AGIQA3K}    & \multicolumn{2}{c}{KonIQ}       & \multicolumn{2}{c}{LIVEC}       & \multicolumn{2}{c}{KADID}       & \multicolumn{2}{c}{LIVE}                             &                \\ \Xhline{1pt}
\multicolumn{1}{c|}{(H,W)}     & PLCC           & SRCC           & PLCC           & SRCC           & PLCC           & SRCC           & PLCC           & SRCC           & PLCC           & SRCC           & PLCC           & SRCC           & PLCC           & SRCC           & PLCC           & SRCC           & PLCC           & SRCC           & PLCC           & SRCC           & PLCC           & SRCC           & PLCC           & \multicolumn{1}{c|}{SRCC}           & Average        \\ \Xhline{1pt}
\multicolumn{1}{c|}{(1,1)}     & 0.920          & 0.912          & \textbf{0.949} & \textbf{0.948} & 0.877          & 0.854          & \textbf{0.908} & \textbf{0.854} & \textbf{0.932} & \textbf{0.913} & \textbf{0.888} & \textbf{0.866} & \textbf{0.880} & \textbf{0.865} & \textbf{0.906} & \textbf{0.852} & \textbf{0.929} & \textbf{0.909} & \textbf{0.902} & \textbf{0.871} & 0.918          & 0.913          & 0.926          & \multicolumn{1}{c|}{0.922}          & \textbf{0.901} \\
\multicolumn{1}{c|}{(7,7)}     & \textbf{0.924} & \textbf{0.917} & 0.947          & 0.948          & 0.876          & 0.852          & 0.906          & 0.844          & 0.931          & 0.912          & 0.881          & 0.848          & 0.875          & 0.857          & 0.902          & 0.843          & 0.928          & 0.906          & 0.890          & 0.855          & \textbf{0.921} & \textbf{0.915} & 0.927          & \multicolumn{1}{c|}{0.931}          & 0.897          \\
\multicolumn{1}{c|}{(14,14)}   & \textbf{0.924} & \textbf{0.917} & 0.947          & 0.948          & 0.876          & 0.852          & 0.906          & 0.844          & 0.931          & 0.912          & 0.881          & 0.848          & 0.875          & 0.857          & 0.902          & 0.843          & 0.928          & 0.906          & 0.890          & 0.855          & \textbf{0.921} & \textbf{0.915} & 0.927          & \multicolumn{1}{c|}{0.931}          & 0.897          \\
\multicolumn{1}{c|}{(28,28)}   & 0.917          & 0.909          & 0.948          & 0.947          & 0.871          & 0.849          & 0.905          & 0.847          & 0.929          & 0.909          & 0.876          & 0.845          & 0.873          & 0.855          & 0.903          & 0.841          & 0.928          & 0.905          & 0.895          & 0.860          & 0.902          & 0.898          & \textbf{0.932} & \multicolumn{1}{c|}{\textbf{0.931}} & 0.895          \\
\multicolumn{1}{c|}{layer\_HW} & 0.921          & 0.914          & 0.948          & 0.948          & \textbf{0.880} & \textbf{0.854} & 0.905          & 0.844          & 0.929          & 0.909          & 0.876          & 0.850          & 0.879          & 0.859          & 0.903          & 0.844          & 0.928          & 0.906          & 0.891          & 0.858          & 0.913          & 0.908          & 0.920          & \multicolumn{1}{c|}{0.919}          & 0.896          \\ \Xhline{2pt}
\end{tabular}

}
}
\caption{Ablation study for different prompt shape}

\label{table: Ablation Size}

\end{table*}

\subsection{Limitations}
\label{Limitations}
Q-Mamba heavily relies on pre-trained weights from ImageNet-1K, which may limit its applicability to domains with significantly different data distributions. Future work could explore pre-training on more diverse datasets to improve generalization capabilities. Despite the efficiency improvements brought by the StylePrompt, the computational demands for training and deploying QMamba on very large-scale datasets or in real-time applications might still be substantial. Investigating methods to further reduce computational complexity without sacrificing performance could be beneficial. The current study focuses on specific types of distortions, and there may be other types of distortions that have not been sufficiently explored. Expanding the evaluation to cover a broader range of distortions could provide a more comprehensive validation of QMamba's robustness. Given that synthetic datasets might not fully capture the complexity and variability of authentic data, there is a risk of overfitting to these synthetic examples. Further evaluations on more diverse and extensive authentic datasets would help ensure the model's robustness and practical applicability. While QMamba shows promising results in cross-domain transferability, its effectiveness across vastly different domains (e.g., medical imaging versus natural images) has not been thoroughly validated. Further studies are needed to test and possibly adapt QMamba for such diverse applications.


\end{document}